\documentclass[letterpaper, 10 pt, conference]{ieeeconf}  

\IEEEoverridecommandlockouts                              

\usepackage{graphics} 
\usepackage{epsfig} 

\usepackage{color}

\usepackage{amssymb}  
\usepackage{amsmath}
\usepackage{epstopdf}
\usepackage{subcaption}
\usepackage{graphicx}
\usepackage{multirow}
\usepackage[table]{xcolor}

\title{\LARGE \bf
Proprioceptive Robot Collision Detection through\\Gaussian Process Regression 
}

\author{Alberto Dalla Libera$^{1}$, Elisa Tosello$^{1}$, Gianluigi Pillonetto$^{1}$, Stefano Ghidoni$^{1}$ and Ruggero Carli$^{1}$
\thanks{$^{1}$A. Dalla Libera, E. Tosello, Gianluigi Pillonetto, S. Ghidoni and R. Carli are with the Deptartment of Information Engineering, University of Padova, Via Gradenigo 6/B, 35131 Padova, Italy
        {\tt\small [dallaliber@dei.unipd.it, toselloe@dei.unipd.it, giapi@dei.unipd.it, ghidoni@dei.unipd.it, carlirug@dei.unipd.it]}}%
}

\begin{document}

\maketitle
\thispagestyle{empty}
\pagestyle{empty}


\begin{abstract}\label{abstract}
This paper proposes a proprioceptive collision detection algorithm based on Gaussian Regression. Compared to sensor-based collision detection and other proprioceptive algorithms, the proposed approach has minimal sensing requirements, since only the currents and the joint configurations are needed. The algorithm extends the standard Gaussian Process models adopted in learning the robot inverse dynamics, using a more rich set of input locations and an ad-hoc kernel structure to model the complex and non-linear behaviors due to frictions in \emph{quasi-static} configurations. Tests performed on a Universal Robots UR10 show the effectiveness of the proposed algorithm to detect when a collision has occurred.

\end{abstract}

\section{Introduction} \label{introduction}

Collaborative Robotics has attracted an increasing interest over the last decade in the research community, mainly due to the fact that the design of robots able to collaborate with humans might have a great impact in several domains.

Human-robot collaboration is a challenging topic under different points of view but, likely, the most critical aspects are related to safety. Indeed, when robots and humans work side-by-side, they need to share their workspace, and, in these circumstances, robots should avoid dangerous and unexpected collision with humans. Despite several motion planning algorithms have been proposed \cite{Pre_collision} in order to minimize the collision probability,
it is impossible to reduce the collision risk to zero. Clearly, in this context it is fundamental that robots are provided with robust strategies that can promptly detect collisions. Moreover, once a collision has been detected, the robot has to classify such collision, in particular discriminating between intended and unintended contacts, and it has to react accordingly.

In order to detect the interaction with the external environment, robots might be endowed with specific sensors, like artificial skins or force-sensors. However, this approach might have some limitations. Indeed artificial skins do not provide information about the collision intensity \cite{Skins}, while six axis force-sensors are expensive and highly sensitive to environmental parameters like temperature and humidity.

A solution alternative to the use of additional sensors is proprioceptive collision detection (CD) \cite{Survey_de_luca}. Proprioceptive collision detection algorithms identify when an external force is applied using only proprioceptive sensors, namely joint torque sensors and current sensors, besides the joint coordinates. We refer the interested reader to  \cite{Survey_de_luca} for an overview of the main state of the art collision detection algorithms. All the proposed approaches require the definition of a monitoring signal $\boldsymbol{s}(t)$ and a threshold $\boldsymbol{\sigma_{CD}}$. The algorithms assume that a collision occurred when $\boldsymbol{s}$ exceeds $\boldsymbol{\sigma_{CD}}$, see \cite{CD_external_torque},\cite{villagrossi}, \cite{CD_power} and \cite{CD_momentum}.
It is worth remarking that these class of solutions require an accurate knowledge of the robot dynamics model, since they assume to know both the kinetics and dynamics parameters. Typically the former parameters are known, while the latter ones are estimated resorting to Fisherian estimators \cite{IdOverview}. 

In this paper, to detect if an interaction has occurred, we propose a novel approach based on the Gaussian Process Regression (GPR) framework. This approach has minimal sensing requirements, since it needs only to measure the joint coordinates and the motor currents.
In this work we extend the GPR algorithms based on semi-parametric priors (i.e., composed by the sum of a parametric component and a non-parametric component) developed to learn the robot inverse dynamics \cite{peters1}, \cite{peters2}, \cite{romeres}. Compared to the standard approach, our algorithm can efficiently deal also with \emph{quasi-static} configurations, namely, when the robot is stuck or the joints' velocities are very low. Specifically, relying on an enlarged set of input features and designing proper kernel structure, our estimator can model the complex behaviors due to static frictions and kinetic frictions at low velocities. 


The paper is organized as follows: in Section \ref{sec:state_of the art} we briefly review state of the art proprioceptive collision detection algorithms based on external torques estimation. In Section \ref{sec:GPRforCD} we present our collision detection strategy, based on Gaussian Regression. In Section \ref{sec:GPR_robotics} we introduce standard GPR techniques adopted in the learning of the robot inverse dynamics, highlighting via a numerical example the limitations of these approaches when used to detect collision in \emph{quasi-static} configurations.  Then, in Section \ref{sec:proposed_approach} we formally describe our learning algorithm and in Section \ref{sec:experiments} we show some numerical results obtained using a UR10 robot.   

\section{CD via monitoring external torques} \label{sec:state_of the art}

In this section we describe a state of the art solution proposed to solve the CD problem, see \cite{Survey_de_luca}.
%
%
When a collision occurs, an external force $\boldsymbol{F_{ext}}(t)$ is applied to the robot, and consequently the joints are subject to a torque $\boldsymbol{\tau_{ext}}(t)$. Consider an $n$ joints manipulator and let $\boldsymbol{q}(t)$, $\boldsymbol{\dot{q}}(t)$, $\boldsymbol{\ddot{q}}(t)$ and $\boldsymbol{\tau_m}(t) \in \mathbb{R}^n$, denote, respectively, the vectors of joints positions, velocities, accelerations and motor torques at time $t$; in the following, to keep the notation compact, we point out explicitly the time dependence only when it is necessary. The expression of $\boldsymbol{\tau_{ext}}$ is given by
	\begin{equation}
	\boldsymbol{\tau_{ext}} = M\left(\boldsymbol{q}\right)\boldsymbol{\ddot{q}} + C\left(\boldsymbol{q}, \boldsymbol{\dot{q}}\right)\boldsymbol{\dot{q}} + \boldsymbol{\tau_g}\left(\boldsymbol{q}\right) + \boldsymbol{\tau_\epsilon} - \boldsymbol{\tau_m} \text{,} \label{eq:tau_ext}
	\end{equation}%
where $ M\left(\boldsymbol{q}\right) \in \mathbb{R}^{n\times n}$ is the generalized inertia matrix, $C\left(\boldsymbol{q}, \boldsymbol{\dot{q}}\right) \in \mathbb{R}^{n\times n}$ is the Coriolis matrix, $\boldsymbol{\tau_g}\left(\boldsymbol{q}\right) \in \mathbb{R}^{n}$ models the effects due to the gravitational force and $\boldsymbol{\tau_\epsilon} \in \mathbb{R}^{n}$ describes the torques related to the unmodeled dynamic behaviors, mainly frictions and elasticity of the links \cite{siciliano}.

Collision detection through direct monitoring of $\boldsymbol{\tau_{ext}}$ defines $\boldsymbol{s}(\boldsymbol{q},\boldsymbol{\dot{q}},\boldsymbol{\ddot{q}}, \boldsymbol{\tau_m})= \boldsymbol{\hat{\tau}_{ext}}(\boldsymbol{q},\boldsymbol{\dot{q}},\boldsymbol{\ddot{q}}, \boldsymbol{\tau_m})$, where $\boldsymbol{\hat{\tau}_{ext}}$ is the estimate of $\boldsymbol{\tau_{ext}}$ obtained from equation \eqref{eq:tau_ext} considering $\boldsymbol{\tau_{\epsilon}}=0$; given measurements of $\boldsymbol{q}$, $\boldsymbol{\dot{q}}$, $\boldsymbol{\ddot{q}}$ and $\boldsymbol{\tau_m}$ we have
\begin{equation}\label{eq:tau_ext_est}
\boldsymbol{\hat{\tau}_{ext}} = M\left(\boldsymbol{q}\right)\boldsymbol{\ddot{q}} + C\left(\boldsymbol{q}, \boldsymbol{\dot{q}}\right)\boldsymbol{\dot{q}} + \boldsymbol{\tau_g}\left(\boldsymbol{q}\right) -\boldsymbol{\tau_m}\text{.}
\end{equation}%
Ideally we should have $\boldsymbol{s}(\cdot)=0$ when $\boldsymbol{\tau_{ext}} = 0$; in practice, given the model inaccuracies and the measurement noise, it happens that the monitoring signal is different from zero even when no external forces are applied. Consequently the introduction of a threshold $\boldsymbol{\sigma_{CD}}$ is necessary, and the binary collision function $f_{CD}(\cdot)$ is defined as
\begin{small}
	\begin{equation*}
	f_{CD}(\boldsymbol{s}) = \begin{cases}
	\text{TRUE, } &if \,\, |\boldsymbol{s}| \geq \boldsymbol{\sigma_{CD}} \\ 
	\text{FALSE, } &if \,\, |\boldsymbol{s}| < \boldsymbol{\sigma_{CD}}
	\end{cases} \text{,}
	\end{equation*}
\end{small}%
where $|\cdot|$, $\geq$ and $<$ are element wise operators, and $|\boldsymbol{s}| \geq \boldsymbol{\sigma_{CD}}$ if the relation holds at least for one component. The value of $\boldsymbol{\sigma_{CD}}$ is set by cross validation with the purpose of limiting the number of false positives and false negatives. Typically the identification of $\boldsymbol{\sigma_{CD}}$ is done observing the evolution of $\boldsymbol{s}(\cdot)$ obtained while the robot is moving with $\boldsymbol{\tau_{ext}}=0$ for a time interval sufficiently large from the statistical point of view, see for examples \cite{Survey_de_luca}.

Finally observe that, in the computation of $\boldsymbol{\hat{\tau}_{ext}}$ in \eqref{eq:tau_ext_est}, it is assumed to know the model of the robotic arm, that is  defined by kinematic parameters and dynamics parameters. Typically kinematic parameters are known while dynamics parameters are estimated by resorting to some Fisherian approach \cite{IdOverview}.

\section{GPR for Proprioceptive collision detection}\label{sec:GPRforCD}

In this paper, we propose a novel approach based on the GPR framework to solve the CD problem. In particular a GPR-based method is used to build the monitoring signal $\boldsymbol{s}$. In the following, instead of measuring directly the torque $\boldsymbol{\tau_m}$, we assume to measure the current $\boldsymbol{i}$ of the motors generating the torque $\boldsymbol{\tau_m}$ applied to the joints; this is due to the fact that in our experimental setup we have access to $\boldsymbol{i}$ and not to $\boldsymbol{\tau_m}$. However, it is worth stressing that a current-based approach has minimal requirements as far as the number of sensors employed is concerned.

To consider the motor currents $\boldsymbol{i}$ instead of $\boldsymbol{\tau_m}$, we need to include the mechanical equations of the motors in the robotic arm model. Let $\boldsymbol{\theta}(t)$, $\boldsymbol{\dot{\theta}}(t)$ and $\boldsymbol{\ddot{\theta}}(t)$ be the angular position, velocity and acceleration of the motors; 
then the mechanical equations of the motors are
	\begin{equation}
	J_m \boldsymbol{\ddot{\theta}} + B_m \boldsymbol{\dot{\theta}} - K_{\tau}\boldsymbol{i} = \boldsymbol{\tau_L} \text{,}\label{eq:dynamic_motor}
	\end{equation}%
where $\boldsymbol{\tau_L}$ are the torques due to the load, and $J_m$, $B_m$ and $K_\tau \in \mathbb{R}^{n \times n}$ are diagonal matrices containing respectively the rotor inertias, the motors damping coefficient and the torques-current ratios. When the behaviors due to the elasticity of the gears are negligible and $\boldsymbol{\tau_{ext}}=0$ it holds $\boldsymbol{\dot{\theta}}=K_r\boldsymbol{\dot{q}}$, $\boldsymbol{\ddot{\theta}}=K_r\boldsymbol{\ddot{q}}$ and $\boldsymbol{\tau_L}=K_r^{-1}\boldsymbol{\tau_m}$, with $K_r \in \mathbb{R}^{n \times n}$ equals to the diagonal matrix containing the gear reduction ratios. Substituting these equations in \eqref{eq:dynamic_motor}, we can express $\boldsymbol{\tau_m}$ as function of $\boldsymbol{q}$,  $\boldsymbol{\dot{q}}$, $\boldsymbol{\ddot{q}}$ and $\boldsymbol{i}$, and equation \eqref{eq:tau_ext} becomes:
	\begin{equation}
	M_{eq}\left(\boldsymbol{q}\right)\boldsymbol{\ddot{q}} + C\left(\boldsymbol{q}, \boldsymbol{\dot{q}}\right)\boldsymbol{\dot{q}} + \boldsymbol{\tau_g}\left(\boldsymbol{q}\right) + \boldsymbol{\tau_\epsilon} + B_{eq} \boldsymbol{\dot{q}} =  K_{eq} \boldsymbol{i} \text{,} \label{eq:tau_ext_curent}
	\end{equation}%
where for compactness we defined $ M_{eq}(\boldsymbol{q}) =  M(\boldsymbol{q}) + K_r^2 J_m$, $B_{eq} = K_r^2B_m$ and $K_{eq} = K_\tau K_r$.

Instead of estimating $\boldsymbol{\tau_{ext}}$ directly from \eqref{eq:tau_ext}, we propose to learn a GPR model that provides an estimate of $\boldsymbol{i}$, denoted as  $\boldsymbol{\hat{i}}$, when $\boldsymbol{\tau_{ext}}$ is null, i.e., $\boldsymbol{\tau_{ext}}=0$; more specifically we train a suitable GPR model for $\boldsymbol{i}$, over a sufficiently rich dataset containing only trajectories obtained with $\boldsymbol{\tau_{ext}}=0$.
Then the monitoring signal $\boldsymbol{s}$ is defined as the difference between the measured current $\boldsymbol{i}$ and $\boldsymbol{\hat{i}}$. Clearly, if no collision has occurred, i.e., $\boldsymbol{\tau_{ext}}$ is effectively null, then we expect $\boldsymbol{\hat{i}}$ to be close to $\boldsymbol{i}$ and, in turn, $\boldsymbol{s}$ to be small; viceversa if a collision has happened, i.e, $\boldsymbol{\tau_{ext}}\neq 0$, then $\boldsymbol{\hat{i}}$ should be significantly different from $\boldsymbol{i}$ and $\boldsymbol{s}$ should become sufficiently large to detect the contact.


We stress the fact that in this paper we focus only on the development of GPR models able to produce proper monitoring signals while we do not discuss any strategy to design the threshold $\boldsymbol{\sigma_{CD}}$. However $\boldsymbol{\sigma_{CD}}$ might be set using standard rules \cite{Survey_de_luca}, like cross-validation.

\section{GPR for robot inverse dynamics}\label{sec:GPR_robotics}

In this Section we briefly introduce the GPR framework \cite{rasmussen}, focusing in the standard models used in the learning of the inverse dynamics, \cite{peters1},\cite{peters2},\cite{romeres}.

Let $\boldsymbol{y}$ be a vector of measurements and let $X = \{\boldsymbol{x_1},\dots, \boldsymbol{x_N}\}$ be the set of the corresponding input locations, with $\boldsymbol{x_k} \in \mathbb{R}^p$, the probabilistic model of GPR is
	\begin{equation}
		\boldsymbol{y} = 
		\begin{bmatrix}
		y_1\\ \vdots \\ y_N
		\end{bmatrix} 
		= 
		\begin{bmatrix}
		f\left(\boldsymbol{x_1}\right)\\ \vdots \\ f\left(\boldsymbol{x_N}\right)
		\end{bmatrix} + 
		\begin{bmatrix}
		e_1\\ \vdots \\ e_N
		\end{bmatrix} 
		= f(X) + e(X) \label{eq:GPR_model}
	\end{equation}%
where $\boldsymbol{e}$ is Gaussian i.i.d. noise with covariance $\sigma_e$ and $f \left( \boldsymbol{x_k} \right): \mathbb{R}^{p}\to \mathbb{R}$ is an unknown function defined as a Gaussian Process, namely $f(X) \sim N\left(\boldsymbol{m_f}(X), K(X,X)\right)$, where $\boldsymbol{m_f}(X)$ is the mean of the process and $K(X,X)$ is the corresponding covariance. Typically $K(X,X)$ is named kernel matrix and it can be defined through a kernel function $k(\boldsymbol{x_i},\boldsymbol{x_j})$, i.e. the $K(X,X)$ entry in $i$-th row and $j$-th column is equal to $k(\boldsymbol{x_i},\boldsymbol{x_j})$. Under these assumptions the posterior probability of $\boldsymbol{f}$ is Gaussian and then $\boldsymbol{\hat{f}}$, the maximum a posteriori estimation of $\boldsymbol{f}$, is the mean of the $\boldsymbol{f}$ posterior distribution. 
	

\subsection{GPR robot inverse dynamics}
The robot inverse dynamics problem consists in learning the function $f$ that maps $\boldsymbol{q}$, $\boldsymbol{\dot{q}}$ and $\boldsymbol{\ddot{q}}$ in $\boldsymbol{\tau_m}$. Typically GPR approaches consider each joint $\ell$ as stand-alone. Referring to the notation introduced in \eqref{eq:GPR_model}, for each joint $\ell$ we introduce a GPR model $y_k= f_\ell(\boldsymbol{x_k}) + e_k$,
where $\boldsymbol{x_k}=\left[\boldsymbol{q}(t_k) , \boldsymbol{\dot{q}}(t_k) , \boldsymbol{\ddot{q}}(t_k) \right]$, $y_k = \tau_{m_\ell}\left(\boldsymbol{x_k}\right)$ and where the $\ell$ subscript denotes that measurement is referred to the $\ell$-th link. Since in our setup we consider currents instead of torques we have $y_k = i_\ell\left(\boldsymbol{x_k}\right)$.

The most crucial aspect in GPR is related to the choice of the prior distribution of $f_\ell(\cdot)$, i.e. the selection of good $m_{f_\ell}(\cdot)$ and $k_\ell(\cdot,\cdot)$. The different priors adopted in Robotics can be grouped in three families, in particular, parametric priors (PPs), non-parametric priors (NPPs) and semi-parametric priors (SPPs).

\subsubsection{Parametric priors}\label{MDP}
When equation \eqref{eq:tau_ext} is given, it is possible to derive an expression of $m_{f_\ell}(\cdot)$ and $k_\ell(\cdot,\cdot)$ which is inspired by the model. Indeed, in \cite{siciliano}, it has been shown that the dynamic model in \eqref{eq:tau_ext} can be rewritten, when neglecting the unmodeled effects, i.e., assuming $\boldsymbol{\tau_\epsilon}=0$, as a linear time-variant model. Formally, when $\boldsymbol{\tau_{ext}}=0$ it holds:
	\begin{align}
	\label{eq:dynamic_linear_model}
	&\boldsymbol{\tau_m} = \begin{bmatrix}
	\boldsymbol{\tau_{m_1}}\\ \vdots \\ \boldsymbol{\tau_{m_n}}
	\end{bmatrix}
	=\begin{bmatrix}
	\boldsymbol{\phi_1^d}\left(\boldsymbol{q},\boldsymbol{\dot{q}},\boldsymbol{\ddot{q}}\right)\\
	\vdots\\
	\boldsymbol{\phi_n^d}\left(\boldsymbol{q},\boldsymbol{\dot{q}},\boldsymbol{\ddot{q}}\right)
	\end{bmatrix}\boldsymbol{w_d} =\Phi^d\left(\boldsymbol{q},\boldsymbol{\dot{q}},\boldsymbol{\ddot{q}}\right)\boldsymbol{w_d} \text{ ,}
	\end{align}%
where $\boldsymbol{w_d} \in \mathbb{R}^m$ denotes the vector casting together all the  dynamic parameters of the robot. The same property holds also if we consider $\boldsymbol{i}$ instead of $\boldsymbol{\tau_m}$, i.e. equation \eqref{eq:tau_ext_curent} instead of \eqref{eq:tau_ext}.

Then, considering $f_\ell(\boldsymbol{x_k}) = \boldsymbol{\phi_\ell^d(\boldsymbol{x_k})}\boldsymbol{w_d}$ with $\boldsymbol{w_d} \sim N(\boldsymbol{m_{w_d}} , \Sigma_{w_d})$, we have
	\begin{equation*}
	\boldsymbol{f}_\ell(X) \sim N\left(\Phi_\ell^d\left(X\right)\boldsymbol{w_d}, \Phi_\ell^d\left(X\right)\Sigma_{w_d}\Phi_\ell^d\left(X\right)^T \right) \text{,}
	\end{equation*}%
with $\Phi_\ell^d\left(X\right) \in \mathbb{R}^{N\times m}$  obtained casting together the vectors $\boldsymbol{\phi}_\ell^d(\cdot)$ evaluated in the input locations of $X$. The kernel function of the process is $k_\ell(\boldsymbol{x_i},\boldsymbol{x_j}) = \boldsymbol{\phi_\ell^d}(\boldsymbol{x_i}) \Sigma_{w_d}\boldsymbol{\phi_\ell^d}(\boldsymbol{x_j})^T$, and it is equivalent to a linear kernel. The mean function is $m_{f_\ell}(\boldsymbol{x_k}) = \boldsymbol{\phi_\ell^d} \left(\boldsymbol{x_k}\right)\boldsymbol{m_{w_d}}$.

A refinement of this model can be obtained including also some terms modeling the frictions effects. The simplest and most used model to describe the torque applied to the $\ell$-th joint by frictions, denoted as $\tau_{f_\ell}$, is given by
\begin{small}
	\begin{equation}
	\label{eq:Friction1}
	\tau_{f_\ell}(t) = \begin{cases}
	\tau_{m_\ell}(t) & \text{if } \dot{q}_{\ell}(t) = 0 \text{ , } \tau_{f_\ell}(t) \le F_{s_\ell}  \\
	F_{k_\ell} sign(\dot{q}_{\ell}(t)) + F_{v_\ell}\dot{q}_{\ell}(t) & \text{if }  |\dot{q}_{\ell}(t)| > 0
	\end{cases} \text{ ,}
	\end{equation}
\end{small}%
where $F_{s_\ell}$, $F_{k_\ell}$ and $ F_{v_\ell}$ are respectively the static friction coefficient, the kinetic friction coefficient and the viscous friction coefficient of the $\ell$-th joint \cite{friction_modeling}. Notice that when $\dot{q}_\ell$ is not null $\tau_{f_\ell}$ is linear respect to  $F_{k_\ell}$ and $F_{v_\ell}$ and hence the behaviors due to the kinetic frictions can be easily merged in \eqref{eq:dynamic_linear_model} leading to the augmented equation
\begin{small}
\begin{equation}\label{eq:Frictions}
	\tau_{m_\ell}(\boldsymbol{x_k}) = \left[ \boldsymbol{\phi_\ell^d} \left(\boldsymbol{x_k}\right) \,\,\, \boldsymbol{\phi_\ell^f} \left(\boldsymbol{x_k}\right)\right] \,
	\left[	
	\begin{array}{c}
	\boldsymbol{w_d} \\
	\boldsymbol{w_{f_\ell}}
	\end{array}
	\right] : = \boldsymbol{\phi_\ell} (\boldsymbol{x_k}) \,  \boldsymbol{w_\ell}
\end{equation}
\end{small}%
where $\boldsymbol{\phi_\ell^f} \left(\boldsymbol{x_k}\right)= \left[sign(\dot{q}_{\ell}) \,\,\,\,\,\dot{q}_{\ell} \right]$ and $\boldsymbol{w_{f_\ell}}=\left[F_{k_\ell}\,\,\,\,\, F_{v_\ell}\right]^T$.


\subsubsection{Non-Parametric priors}\label{DAP}

When no prior knowledge about the process is available, the most common choice is to consider $m_{f_\ell}(\cdot) = 0$ and define $K_\ell(X,X)$ directly through a kernel function $k_\ell(\cdot,\cdot)$. The most used kernel in robotic identification is the Radial Basis Kernel (RBK), defined as
\begin{small}
	\begin{equation}
	\label{eq:Gaussian_kernel}
	k_{RBK}(\boldsymbol{x_i}\boldsymbol{x_j}) = \lambda \exp\left(-\frac{\left(\boldsymbol{x_i}-\boldsymbol{x_j}\right)^T \Sigma_{RBK}^{-1} \left(\boldsymbol{x_i}-\boldsymbol{x_j}\right)}{2}\right)
	\end{equation}
\end{small}%
where $\Sigma_{RBK}$ is typically a diagonal matrix, whose diagonal elements $\sigma_{RBK}$ are referred to as length-scales. 


\subsubsection{Semi-Parametric priors}\label{HP}

The semi-parametric approach models the function $f_\ell(\cdot)$ as the sum of two independent contributions, a parametric component $f_{P_\ell}= \boldsymbol{\phi_\ell^d}(\boldsymbol{x_k})\boldsymbol{w_d}$ and a non-parametric component $f_{NP_\ell}(\cdot)$, for example defined by an RBK i.e. $f_\ell(\boldsymbol{x_k}) = \boldsymbol{\phi_\ell^d}(\boldsymbol{x_k}) \boldsymbol{w_d} + f_{NP_\ell}(\boldsymbol{x_k})$. Typically, there are two ways to include the parametric component. $(i)$ Assuming that $\boldsymbol{w_d}$ is a deterministic variable, eventually pre-trained adopting a parametric-based estimator; in this case the mean and the kernel of $f_\ell(\cdot)$ is $m_{f_\ell}(\boldsymbol{x_k})= \boldsymbol{\phi_\ell^d}(\boldsymbol{x_k})\boldsymbol{w_d}$ and $k_\ell(\boldsymbol{x_i}, \boldsymbol{x_j}) = k_{RBK}(\boldsymbol{x_i},\boldsymbol{x_j})$. $(ii)$ Assuming that $\boldsymbol{w_d}$ is a random variable independent from $f_{NP_\ell}(\cdot)$, thus obtaining $m_{f_\ell}(\boldsymbol{x_k})= \boldsymbol{\phi_\ell^d}(\boldsymbol{x_k})\boldsymbol{m_{w_d}}$ and $k_\ell(\boldsymbol{x_i}, \boldsymbol{x_j}) = \boldsymbol{\phi_\ell^d}(\boldsymbol{x_i}) \Sigma_w\boldsymbol{\phi_\ell^d}(\boldsymbol{x_j})^T + k_{RBK}(\boldsymbol{x_i},\boldsymbol{x_j})$.


\subsection{Limitations of proprioceptive collision detection with standard GPR approach} \label{sec:standard_GPR_CD}

\begin{figure}
	\centering
	\includegraphics[width=\linewidth]{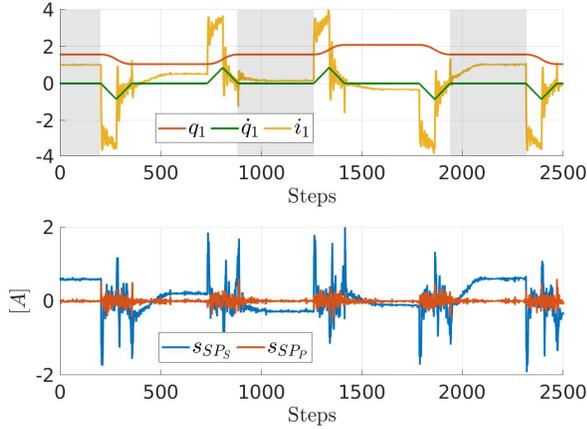}
	\caption{Cyclic actuation of the first link.  $S_{SP_S}$ and $SP_P$ denote the monitoring signals obtained, respectively, by a standard semi-parametric estimator and the proposed approach.}
	\label{fig:joint1movement}
\end{figure}

In this subsection we discuss a simple experiment that highlights the limitations of these standard GPR estimators when working in the \emph{quasi-static} configurations. The experiment, reported in Figure \ref{fig:joint1movement},  consists in a succession of rest phases (all the joints stuck and parallel to the ground) and moving phases. In the moving phases only the first joint is actuated, such that the values of $q_1$ in the rest phase are sequentially $\frac{\pi}{2}$, $2.09$, $\frac{\pi}{2}$ , $0.52$ $[rad]$.

In Figure \ref{fig:joint1movement}, the blue line represents the monitoring signal $s_{SP_S}$ obtained estimating the current using a standard semi-parametric estimator. Notice that, while during the moving phases the frequency of $s_{SP_S}$ is particularly high and it might be easily canceled with a low pass filter, during the rest phases $s_{SP_S}$ is significantly greater than zero for sufficiently long intervals. Consequently a collision might be detected, generating a false positive. 

This fact is caused by the poor estimation performances of the standard GPR estimators when the robot is in \emph{quasi-static} configurations (see results in Section \ref{sec:exp_rand_exploration}). Indeed at low velocities the forces due to frictions are more relevant and particularly unpredictable \cite{friction_modeling}. As confirmed by equation \eqref{eq:Frictions}, when $|\dot{q_\ell}|<\sigma_v$ the model is highly non linear and strongly dependent on different factors like the physical properties of the materials. The threshold $\sigma_v$ defines the transition between \emph{dynamical} and \emph{quasi-static} configurations. Its value can be validated via cross-correlation and in this paper it has been set equal to $10^{-2}$. See \cite{giappo} for details.


This experiment shows another interesting fact explaining the reason why the  non-parametric component is not able to capture the behaviors due to $\boldsymbol{\tau_f}$ when $|\dot{q}|<\sigma_v$. Observe that in the rest phases with $q_1 = \pi/2$, despite the robot is in the same configuration $\boldsymbol{x_{q}}$, the current $i_1$ assumes three different values. Referring to the GPR notation, the function $f_1(\cdot)$ attains different values in the same input location $\boldsymbol{x_{q}}$, and the difference among these values is so significant that can not be explained by only the presence of noise in the measurements. Similar situation happens in linear classification, when two classes are not linearly separable, and it denotes the need of more input features.

\section{Proposed learning algorithm} \label{sec:proposed_approach}

The proposed solution is based on the following observations. $(i)$ Experimental results in Section \ref{sec:exp_rand_exploration} show that, when working in a \emph{dynamical} configuration, a semi-parametric kernel provides accurate estimates when describing the input locations by the standard features $\boldsymbol{q}$, $\boldsymbol{\dot{q}}$, $\boldsymbol{\ddot{q}}$. $(ii)$ When dealing with the \emph{quasi-static} configuration, we need to include additional features in the input space, in order to avoid that the same input is mapped into different output. $(iii)$ We need to model the discontinuity due to the different behaviors of static frictions and kinetic frictions, i.e., we need to provide a unified framework capturing the behaviors in both scenarios, \emph{dynamical} and \emph{quasi-static}.  

Based on the above observations the learning algorithm we propose models the function $f_\ell$ adopting a semi-parametric model, where the parametric component $f_{P_\ell}$ includes also the frictions effects and where the non-parametric component $f_{NP_\ell}$ is given by the sum of two contributions; the first one trying to compensate the model inaccuracies and the second one capturing the discontinuous behaviors generated by the frictions in the transition intervals between static and dynamic frictions. 

Before formally describing the model we consider, we provide some more details about the second and third observation above.

\subsection{Additional features}

%
Notice, from Equation \eqref{eq:Friction1},  that when the velocity is null, important contributions are given by $\boldsymbol{\tau_m}$, that is a term related to the action of the controller. Consequently in \emph{quasi-static} configurations it might be necessary to add to the GPR inputs some features related to the control actions. We stress the fact that, from a control point of view, we are operating in a black box context since we do not have access to the low level controller of the UR robot we used in our experiments.

In our learning algorithm, when dealing with the \emph{quasi-static} case, the input locations are described by the following augmented features vector,
	\begin{equation}
	\boldsymbol{x_k^{a}} = \begin{bmatrix}
	\boldsymbol{q}(t_k), \boldsymbol{\dot{q}}(t_k), \boldsymbol{\ddot{q}}(t_k), \boldsymbol{e_q}(t_k), \boldsymbol{\dot{e}_q}(t_k), \boldsymbol{i_c}(t_k)
	\end{bmatrix} \label{eq:augmented_inputs}
	\end{equation}
where $\boldsymbol{e_q}(t_k)$ and $\boldsymbol{\dot{e}_q}(t_k)$ denote, respectively, the joint position and velocity errors at time $t_k$, while $\boldsymbol{i_c}(t_k)$ are the currents required by the controllers of the motors at the instant $t_k$.

The rationale behind the choice of adopting this set of features is the following:
the variables $\boldsymbol{e_q}$ and $\boldsymbol{\dot{e}_q}$ allow to model proportional and derivative contributions while the $\boldsymbol{i_c}$  bring information about non linear control actions (i.e. saturation) and dynamic contribution (i.e. integral contribution).

\subsection{Modeling of friction discontinuity through NPP}

In our approach, to capture the discontinuity between static frictions and kinematic frictions we add to our model a non-parametric component. In the following, to keep compact the description of our model, we exploit two properties of kernels functions \cite{rasmussen}. $(i)$ The sum of kernels is a kernel. $(ii)$ Vertical rescaling: let $k(\cdot, \cdot)$ be the kernel function of the Gaussian Process $f(\boldsymbol{x_k})$ and $a(\boldsymbol{x_k})$ a deterministic function. Then $a(\boldsymbol{x_i})k(\boldsymbol{x_i},\boldsymbol{x_j})a(\boldsymbol{x_j})$ is a valid kernel function and in particular it is associated to the process $a(\boldsymbol{x_k})f(\boldsymbol{x_k})$.

The non-parametric component $f_{NP_\ell}$ of our model is given as the sum of two Gaussian Processes, $f_{NP_{\ell; stc}}(\cdot)$ and $f_{NP_{\ell; kin}}(\cdot)$, where the first one is scaled by the function 
	\begin{equation*}
	a_\ell(\boldsymbol{x_k^{a}}) = \begin{cases}
	0 & if \,\,|\dot{q}_{k_\ell}|\geq \sigma_v\\
	1 & if \,\,|\dot{q}_{k_\ell}|<\sigma_v
	\end{cases} \text{.}
	\end{equation*}%
It turns out that
\begin{small}
\begin{equation}
f_{NP_\ell}(\boldsymbol{x_k^{a}}) = a_\ell(\boldsymbol{x_k^{a}})f_{NP_{\ell; stc}}(\boldsymbol{x_k^{a}}) + f_{NP_{\ell; kin}}(\boldsymbol{x_k}) \text{.}\label{eq:DDP_proposed}
\end{equation}
\end{small}%

The first component (that is a function of the augmented input vector $\boldsymbol{x_k^a}$), acts only when the $\ell$-th link is in \emph{quasi-static} configurations, with the specific task of capturing the behaviors due to frictions at low velocity. Instead $f_{NP_{\ell;kin}}$ tries to compensate for the PP inaccuracies, and it is active on both the \emph{dynamical} and \emph{quasi-static} configurations; for this reason it depends only on $\boldsymbol{q}, \boldsymbol{\dot{q}}, \boldsymbol{\ddot{q}}$ and not on the additional features $\boldsymbol{e_q}, \boldsymbol{\dot{e}_q}, \boldsymbol{i_c}$\footnote{Formally, in \eqref{eq:DDP_proposed}, $f_{NP_{\ell; kin}}$ should depend on $\boldsymbol{x_k^{a}}$. However, based on the observation reported, we have made explicit the fact that the additional features do not affect the value of $f_{NP_{\ell; kin}}$ which depends only on the standard features $\boldsymbol{x_k}=\left(\boldsymbol{q}(t_k), \boldsymbol{\dot{q}}(t_k), \boldsymbol{\ddot{q}}(t_k)\right)$.}

\subsection{Proposed Algorithm}\label{approach}
The proposed learning algorithm is based on a semi-parametric model described by the following expression
	\begin{equation}
	f_\ell(\boldsymbol{x_k^{a}}) = \boldsymbol{\bar{\phi}_\ell}(\boldsymbol{x_k}) \boldsymbol{w_\ell} +  a_\ell(\boldsymbol{x_k})f_{NP_{\ell; stc}}(\boldsymbol{x_k^{a}}) + f_{NP_{\ell; kin}}(\boldsymbol{x_k}) \text{,} \label{eq:proposed_apporach}
	\end{equation}%
where $\boldsymbol{\bar{\phi}_\ell}(\cdot)$ is defined as $\boldsymbol{\phi_\ell}(\cdot)$, except that  the contributions of $\boldsymbol{\phi_\ell^f}(\cdot)$ are nulled when $|\dot{q}_\ell|<\sigma_v$. This choice is motivated by the experimental evidence that shows how the linear model is not accurate in \emph{quasi-static} configurations.

In our implementation the information coming from the parametric contribution is added considering $\boldsymbol{w_\ell}$ as a deterministic value, i.e. influencing only the mean of $f_\ell(\cdot)$. As far as the $f_{NP_{\ell; stc}}$ and $f_{NP_{\ell; kin}}$ components are concerned, we defined them adopting RBK kernels with ARD.

\section{EXPERIMENTS}\label{sec:experiments}

A Universal Robots UR10\footnote{www.universal-robots.com/UR10} is used for the experiments. It is a collaborative industrial robot with 6-degrees of freedom. The interface with the UR10 is based on ROS (Robot Operating System, \cite{ROS}), through the \textit{ur\_modern\_driver}\footnote{https://github.com/ThomasTimm/ur\_modern\_driver}. Data are acquired with a sampling time of $8\cdot 10^{-3}sec$. The data processing and the derivation of the physical model are implemented in MATLAB, while the GPR in Python, in order to exploit the PyTorch computational advantages during the model optimization \cite{pytorch}.

The normalized mean squared error (nMSE) between $\boldsymbol{i}$ and $\boldsymbol{\hat{i}}$ has been considered in order to evaluate the algorithms accuracy,
\begin{equation}
nMSE(X) = \frac{ \sum_{k = 1}^{N}( i_\ell(\boldsymbol{x_k}) - \hat{i}_\ell(\boldsymbol{x_k}))^2/N}{ Var\left(\boldsymbol{i_\ell}(X_*)\right)} \text{,} \label{eq:nMSE}\\
\end{equation}%

The algorithms tested are $P_f$, a PP-based estimator with linear features modeling kinetic frictions, $SP_S$, a SPP-based estimator with standard input features and $SP_P$, the proposed approach.

\subsection{Random exploration of the workspace} \label{sec:exp_rand_exploration}
In this experiment we test the estimation performances of the learning algorithms, stressing the generalization properties. We considered two data sets. The first is pointed by $D_1$, and it consist in a set of trajectories collected requiring to the end-effector to reach 200 random points (for a total of 80000 input locations) randomly distributed within an hemisphere of the robot workspace. The other data set $D_2$ is composed by 22000 data points collected requiring the robot to reach 50 random points inside the previous hemisphere and to track a circle of radius $30[cm]$ at a tool speed of $30[mm/s]$.
The algorithms have been trained minimizing the negative marginal log likelihood (MLL) over $D_1$. Given the number of samples, to minimize the negative MLL we resorted to stochastic gradient descent \cite{hinton}, in particular we adopted the ADAM optimizer \cite{ADAM1}. Furthermore, once the hyperparameters have been selected, we down-sampled the training set to obtain $D_{SDP}$, a subset of data points with $5000$ samples, used to derive the estimation; the set composed by the remaining input locations is $D_{1_{test}}$. The performances over $D_{1_{test}}$ compare the estimators accuracy in points that are close to $D_{SDP}$, i.e. in points that are close to the input locations used to derive the model. In contrast $D_2$ is thought to stress the estimators generalization properties, and might contains input locations that are far from the ones in $D_{SDP}$.



\begin{figure}
	\centering
	\includegraphics[width=\linewidth]{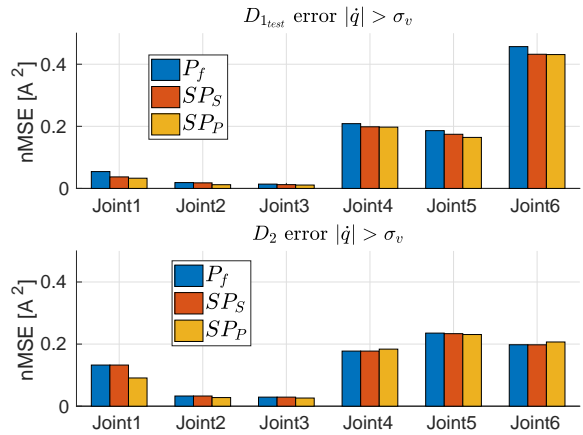}
	\caption{Bar-plot of the nMSE in \emph{dynamical} configurations.
	}
	\label{fig:dynamic_test}
\end{figure}

Results reported in Figure \ref{fig:dynamic_test} show that when links are in non static configurations, namely, when $|\dot{q}_\ell|>\sigma_v$, performances of all the estimators are comparable. This is related to the fact that in these configurations the parametric contributions can capture a relevant part of the signal.

Comparing the nMSE of $SP_S$ and $P_f$, we can appreciate that the addition of the NP contribution in $SP_S$ allows to improve the accuracy in points that are close to $D_{1_{SDP}}$, since $SP_S$ over-performs $P_f$ in $D_{1_{test}}$. However the NP contribution tends to vanish when $SP_S$ is tested in $D_2$. 

In \emph{dynamical} configurations the proposed approach behaves similarly to $SP_S$. However notice that in joint 1 and 2 $SP_P$ significantly improves the $P_f$ performances, even in $D_2$. This aspect suggests that the ad-hoc kernel structure proposed in \eqref{eq:DDP_proposed} entails advantages also in the \emph{dynamical} configurations.

\begin{figure}
	\centering
	\includegraphics[width=\linewidth]{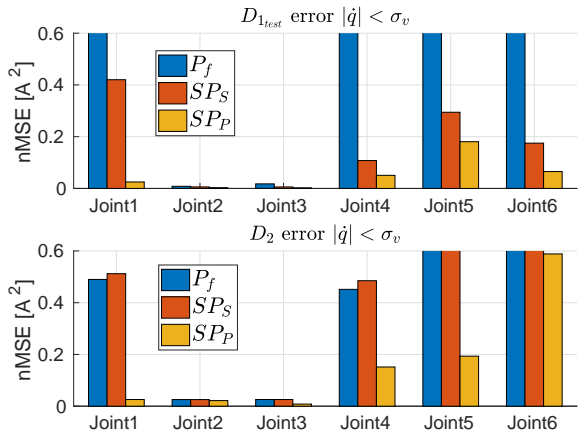}
	\caption{Bar-plot of the nMSE in \emph{quasi-static} configurations.
	}
	\label{fig:static_test}
\end{figure}

The nMSEs in the static and \emph{quasi-static} configurations are reported in Figure \ref{fig:static_test}. The bar-plot highlights that in these configurations $P_f$ does not capture relevant components of the output signal, except for joint 2 and 3. Indeed in link 2 and 3 when the robot is in static configurations the gravitational contributions are predominant, and $P_f$ is able to capture them.

The nMSE index for $SP_S$ highlights how the NP contribution in $SP_S$ is extremely local, since it reduces considerably the nMSE only over $D_{1_{test}}$. Moreover, the $SP_S$ performance over $D_2$ suggests that the semi-parametric estimator with standard inputs is subject to overfitting, give that its nMSE is greater that the one of $P_f$.

The $SP_P$ estimator instead exhibits good performances in static and \emph{quasi-static} configurations over both the datasets, suggesting that the additional features, together with the ad-hoc kernel structure are crucial to model the complex behaviors generated by static frictions.

\subsection{Detection of human-robot interaction}
\begin{figure}
	\centering
	\includegraphics[width=\linewidth]{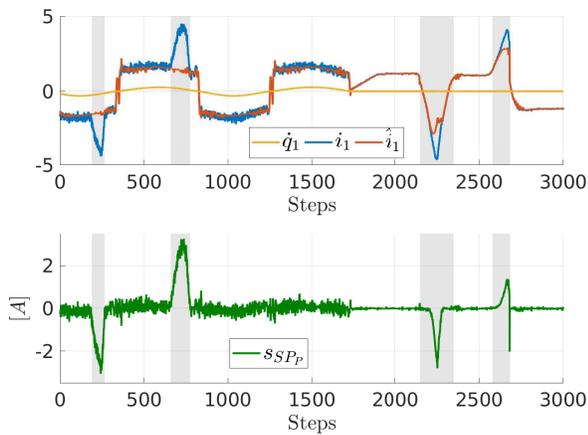}
	\caption{Evolution of $i_1$, $\hat{i}_1$, $\dot{q}_1$ and $s_{SP_{P}}$ when external forces are applied to the fist link of the UR10. The gray bars indicates the interval in which the interactions occurred.}
	\label{fig:external_load}
\end{figure}

In order to validate the CD algorithm proposed, we applied $SP_P$ on a real test case: the detection of human-robot interaction. We tested the algorithm both in \emph{dynamical} and \emph{quasi-static} configurations. In the first part of the experiment the end-effector of the robot is tracking a circle, while in the last part it stays in the final configuration. A human user applies an external force to the first robot joint four times, two during the moving phase and two during the \emph{quasi-static} phase\footnote{The experiment is visible at https://youtu.be/2jJS8ajXhEw}. The $SP_P$ estimator is the same of Experiment \ref{sec:exp_rand_exploration}, derived starting from the $D_{SDP}$ data. The experiment is described in Figure \ref{fig:external_load}. The gray bar highlights the time intervals in which the interactions occurred. The results show that $SP_P$ can be exploited to define a good monitoring signal. Indeed the prediction error $s_{SP_{P}}$ is significantly not null only when the external forces are applied, allowing the detection of the interactions, and at the same time avoiding the possibility of incurring in false positives.

\section{CONCLUSIONS} \label{sec:conclusion}

In this paper we validated the use of GPR to solve the proprioceptive collision detection problem, focusing on the definition of a good monitoring signal. The proposed approach has minimal requirements in terms of sensors, since only joint coordinates and motor currents are needed. The proposed monitoring signal corresponds to the estimate of the currents due to external torques.
In particular we focused on the behaviors of the monitoring signal in static and \emph{quasi-static} configurations, that are particularly relevant in collaborative robotics. 
The proposed approach has been tested in a UR10. The experimental results prove the validity of these methods.


\addtolength{\textheight}{-12cm}   

\bibliographystyle{IEEEtran}
\bibliography{references}

\begin{thebibliography}{10}
\providecommand{\url}[1]{#1}
\csname url@samestyle\endcsname
\providecommand{\newblock}{\relax}
\providecommand{\bibinfo}[2]{#2}
\providecommand{\BIBentrySTDinterwordspacing}{\spaceskip=0pt\relax}
\providecommand{\BIBentryALTinterwordstretchfactor}{4}
\providecommand{\BIBentryALTinterwordspacing}{\spaceskip=\fontdimen2\font plus
\BIBentryALTinterwordstretchfactor\fontdimen3\font minus
  \fontdimen4\font\relax}
\providecommand{\BIBforeignlanguage}[2]{{%
\expandafter\ifx\csname l@#1\endcsname\relax
\typeout{** WARNING: IEEEtran.bst: No hyphenation pattern has been}%
\typeout{** loaded for the language `#1'. Using the pattern for}%
\typeout{** the default language instead.}%
\else
\language=\csname l@#1\endcsname
\fi
#2}}
\providecommand{\BIBdecl}{\relax}
\BIBdecl

\bibitem{Rehabilitation}
G.~J. Gelderblom, M.~D. Wilt, G.~Cremers, and A.~Rensma, ``Rehabilitation
  robotics in robotics for healthcare; a roadmap study for the european
  commission,'' in \emph{2009 IEEE International Conference on Rehabilitation
  Robotics}, June 2009, pp. 834--838.

\bibitem{Collaborative}
P.~Masinga, H.~Campbell, and J.~A. Trimble, ``A framework for human
  collaborative robots, operations in south african automotive industry,'' in
  \emph{2015 IEEE International Conference on Industrial Engineering and
  Engineering Management (IEEM)}, Dec 2015, pp. 1494--1497.

\bibitem{Pre_collision}
D.~M. Ebert and D.~D. Henrich, ``Safe human-robot-cooperation: image-based
  collision detection for industrial robots,'' in \emph{IEEE/RSJ International
  Conference on Intelligent Robots and Systems}, vol.~2, Sept 2002, pp.
  1826--1831 vol.2.

\bibitem{Skins}
A.~Cirillo, F.~Ficuciello, C.~Natale, S.~Pirozzi, and L.~Villani, ``A
  conformable force/tactile skin for physical human–robot interaction,''
  \emph{IEEE Robotics and Automation Letters}, vol.~1, no.~1, pp. 41--48, Jan
  2016.

\bibitem{Survey_de_luca}
S.~Haddadin, A.~D. Luca, and A.~Albu-Schäffer, ``Robot collisions: A survey on
  detection, isolation, and identification,'' \emph{IEEE Transactions on
  Robotics}, vol.~33, no.~6, pp. 1292--1312, Dec 2017.

\bibitem{CD_external_torque}
D.~P. Le, J.~Choi, and S.~Kang, ``External force estimation using joint torque
  sensors and its application to impedance control of a robot manipulator,'' in
  \emph{2013 13th International Conference on Control, Automation and Systems
  (ICCAS 2013)}, Oct 2013, pp. 1794--1798.

\bibitem{villagrossi}
E.~Villagrossi, ``Robot dynamic modelling and control for machining
  applications,'' Ph.D. dissertation, University degli Studi di Brescia, 2015.

\bibitem{CD_power}
A.~D. Luca, A.~Albu-Schaffer, S.~Haddadin, and G.~Hirzinger, ``Collision
  detection and safe reaction with the dlr-iii lightweight manipulator arm,''
  in \emph{2006 IEEE/RSJ International Conference on Intelligent Robots and
  Systems}, Oct 2006, pp. 1623--1630.

\bibitem{CD_momentum}
G.~Doisy, ``Sensorless collision detection and control by physical interaction
  for wheeled mobile robots,'' in \emph{2012 7th ACM/IEEE International
  Conference on Human-Robot Interaction (HRI)}, March 2012, pp. 121--122.

\bibitem{IdOverview}
J.~Wu, J.~Wang, and Z.~You, ``An overview of dynamic parameter identification
  of robots,'' \emph{Robotics and Computer-Integrated Manufacturing}, vol.~26,
  no.~5, pp. 414 -- 419, 2010.

\bibitem{peters1}
J.~Nakanishi, R.~Cory, M.~Mistry, J.~Peters, and S.~Schaal, ``Operational space
  control: A theoretical and empirical comparison,'' \emph{International
  Journal of Robotics Research}, vol.~27, no.~6, pp. 737--757, Jun. 2008.

\bibitem{peters2}
J.-A. Ting, M.~Mistry, J.~Peters, S.~Schaal, and J.~Nakanishi, ``A bayesian
  approach to nonlinear parameter identification for rigid body dynamics,'' in
  \emph{RSS 2006}, Max-Planck-Gesellschaft.\hskip 1em plus 0.5em minus
  0.4em\relax Cambridge, MA, USA: MIT Press, Apr. 2007, pp. 247--254.

\bibitem{romeres}
D.~Romeres, M.~Zorzi, R.~Camoriano, and A.~Chiuso, ``Online semi-parametric
  learning for inverse dynamics modeling,'' in \emph{2016 IEEE 55th Conference
  on Decision and Control (CDC)}, 2016.

\bibitem{siciliano}
B.~Siciliano, L.~Sciavicco, L.~Villani, and G.~Oriolo, \emph{Robotics,
  Modelling, Planning and Control}, 2009.

\bibitem{rasmussen}
C.~E. Rasmussen, ``Gaussian processes for machine learning.''\hskip 1em plus
  0.5em minus 0.4em\relax MIT Press, 2006.

\bibitem{friction_modeling}
P.~E. Dupont, ``Friction modeling in dynamic robot simulation,'' in
  \emph{Robotics and Automation}, 1990.

\bibitem{giappo}
N.~D. Vuong and M.~H.~A. Jr, ``Dynamic model identification for industrial
  robots,'' in \emph{IEEE Control Systems}, 2007.

\bibitem{ROS}
M.~Quigley, K.~Conley, B.~P. Gerkey, J.~Faust, T.~Foote, J.~Leibs, R.~Wheeler,
  and A.~Y. Ng, ``Ros: an open-source robot operating system,'' in \emph{ICRA
  Workshop on Open Source Software}, 2009.

\bibitem{pytorch}
A.~Paszke, S.~Gross, S.~Chintala, G.~Chanan, E.~Yang, Z.~DeVito, Z.~Lin,
  A.~Desmaison, L.~Antiga, and A.~Lerer, ``Automatic differentiation in
  pytorch,'' 2017.

\bibitem{hinton}
G.~E. Hinton, N.~Srivastava, A.~Krizhevsky, I.~Sutskever, and R.~R.
  Salakhutdinov, ``Improving neural networks by preventing co-adaptation of
  feature detectors,'' in \emph{https://arxiv.org/abs/1207.0580}, 2012.

\bibitem{ADAM1}
D.~P. Kingma and J.~L. Ba, ``Adam : A method for stochastic optimization,'' in
  \emph{ICLR}, 2015.

\end{thebibliography}

\end{document}